\documentclass[10pt,twocolumn,letterpaper]{article}

\usepackage{wacv}
\usepackage{times}
\usepackage{epsfig}
\usepackage{amsmath}
\usepackage{amssymb}
\usepackage{subfig}
\usepackage{array}
\usepackage{graphicx,xcolor}
\usepackage[normalem]{ulem}
\usepackage{arydshln}
\usepackage{caption}
\usepackage{enumitem}

\newcommand{\ignore}[1]{}

\newcommand{\bh}{{\mathbf {h}}}

\newcommand{\bp}{{\mathbf {p}}}

\newcommand{\bx}{{\mathbf {x}}}
\newcommand{\bA}{{\mathbf {A}}}
\newcommand{\bI}{{\mathbf {I}}}
\newcommand{\bY}{{\mathbf {Y}}}
\newcommand{\bC}{{\mathbf {C}}}

\newcommand{\PLACES}{{ {Places205}}}

\newcommand{\mTABLE}[1]{Table~#1}
\newcommand{\mFIG}[1]{Fig.~#1}
\newcommand\circled[1][red]{%
    \raisebox{-0.9ex}{\scalebox{3}{\textcolor{#1}{\textbullet}}}
  }%
\newcommand{\textcolorM}[1]{\textcolor{black}}

\definecolor{anger}{rgb}{1, 0, 0}
\definecolor{disgust}{rgb}{1, 0.7529, 0}
\definecolor{fear}{rgb}{0, 0.4392, 0.7529}
\definecolor{joy}{rgb}{0.5725, 0.8156, 0.3137}
\definecolor{sadness}{rgb}{0.8509, 0.588,0.580}
\definecolor{surprise}{rgb}{0, 0.690, 0.3137}
\definecolor{neutral}{rgb}{0.4392, 0.188, 0.62745}

\wacvfinalcopy 

\def\wacvPaperID{239} 

\ifwacvfinal\fi
\begin{document}

\title{High-Level Concepts for Affective Understanding of Images}


\author{Afsheen Rafaqat Ali \\
Information Technology\\
University of Punjab, Lahore\\
{\tt\small afsheen.ali@itu.edu.pk}
\and
Usman Shahid \\
Information Technology\\
University of Punjab, Lahore\\
{\tt\small usman.shaahid@gmail.com }
\and
Mohsen Ali \\
Information Technology\\
University of Punjab, Lahore\\
{\tt\small mohsen.ali@itu.edu.pk}
\and
Jeffrey Ho \\
{\tt\small jho.jeffrey@gmail.com}
}

\maketitle
\ifwacvfinal\thispagestyle{empty}\fi

\begin{abstract}
This paper aims to bridge the affective gap between image content and the emotional response of the viewer it elicits by using High-Level Concepts (HLCs). In contrast to previous work that relied solely on low-level features or used convolutional neural network (CNN) as a black-box, we use HLCs generated by pretrained CNNs in an explicit way to investigate the relations/associations between these HLCs and a (small) set of Ekman's emotional classes. As a proof-of-concept, we first propose a linear admixture model for modeling these relations, and the resulting computational framework allows us to determine the associations between each emotion class and certain HLCs (objects and places). This linear model is further extended to a nonlinear model using support vector regression (SVR) that aims to predict the viewer's emotional response using both low-level image features and HLCs extracted from images. These class-specific regressors are then assembled into a regressor ensemble that provide a flexible and effective predictor for predicting viewer's emotional responses from images. Experimental results have demonstrated that our results are comparable to existing methods, with a clear view of the association between HLCs and emotional classes that is ostensibly missing in most existing work.
\end{abstract}

\begin{table}
\center
\begin{tabular}{ m{3.4em} : m{4.3cm}   m{1.9cm}}
\small{High-Level Concept}& \center\small{Related images from Emotion6 dataset} & \small{Mean emotion distribution and dominant emotion class} \\ \hline
  \small{Butterfly/ Flowers}  & \subfloat{ \includegraphics[width=1.45cm, height=1.3cm]{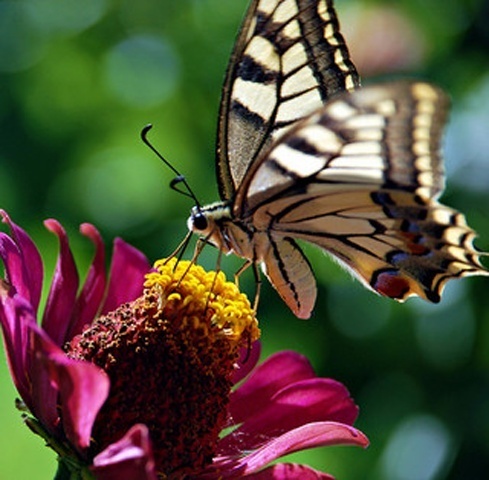}} 
  \subfloat{
  \includegraphics[width=1.45cm, height=1.3cm]{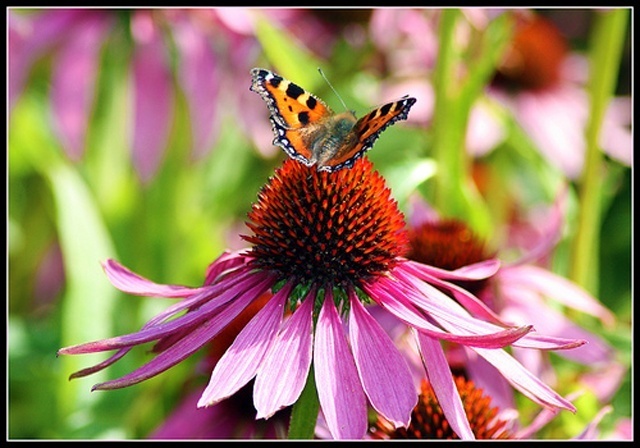}}
  \subfloat{ \includegraphics[width=1.45cm, height=1.3cm]{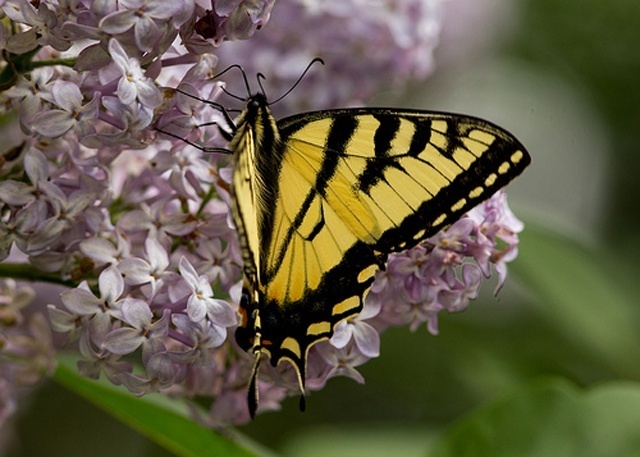}} &
  \subfloat{ \includegraphics[width=1.6cm, height=1.3cm]{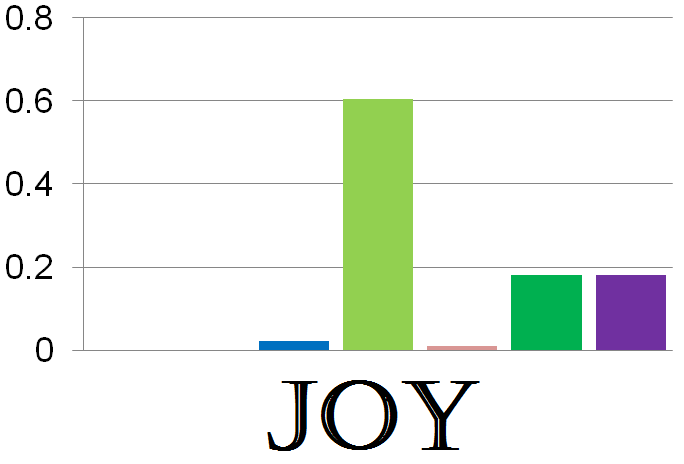}}\\ [-1.9ex]
    \small{Toilet Seat }  & \subfloat{ \includegraphics[width=1.45cm, height=1.3cm]{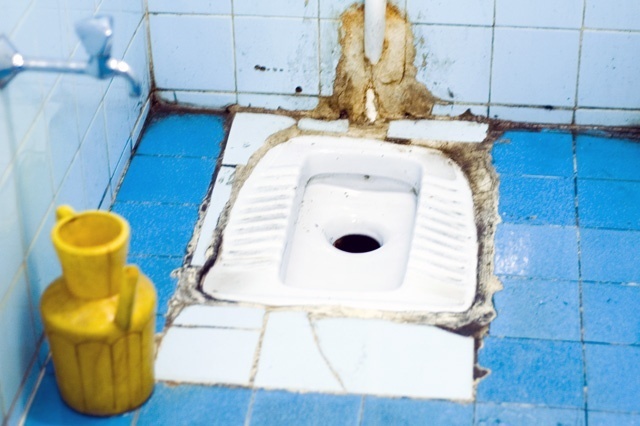}} 
  \subfloat{
  \includegraphics[width=1.45cm, height=1.3cm]{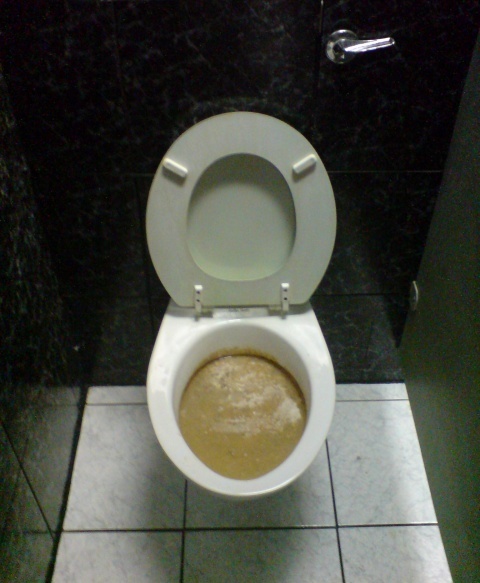}}
  \subfloat{ \includegraphics[width=1.45cm, height=1.3cm]{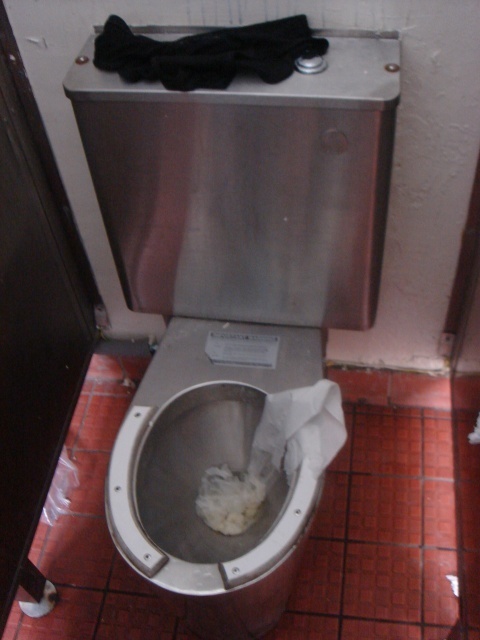}} & \subfloat{ \includegraphics[width=1.6cm, height=1.3cm]{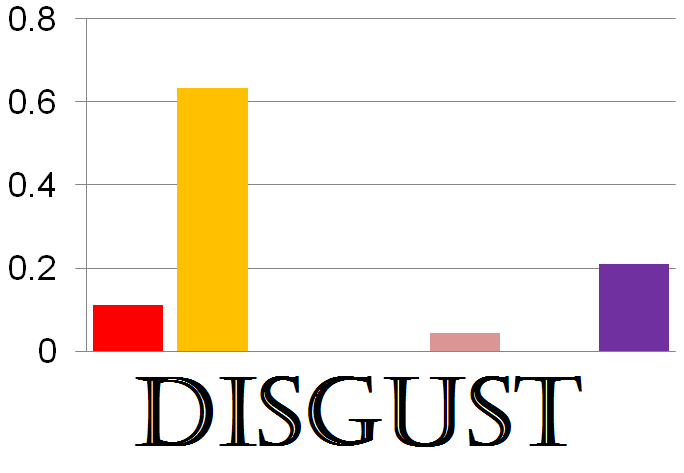}} \\ [-1.9ex]
      \vspace{-0.5cm}
    \small{Mask } &\subfloat{ \includegraphics[width=1.45cm, height=1.3cm]{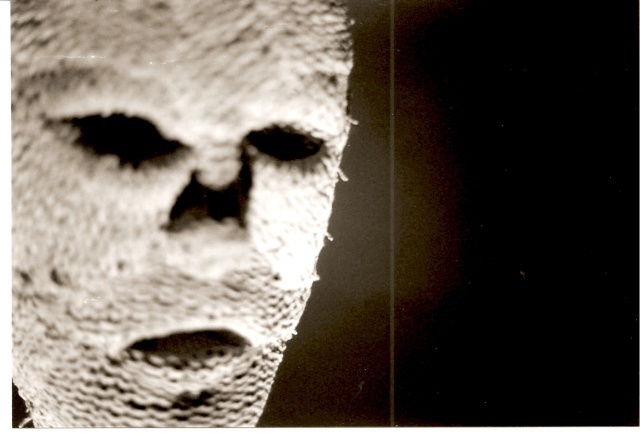}} 
  \subfloat{
  \includegraphics[width=1.45cm, height=1.3cm]{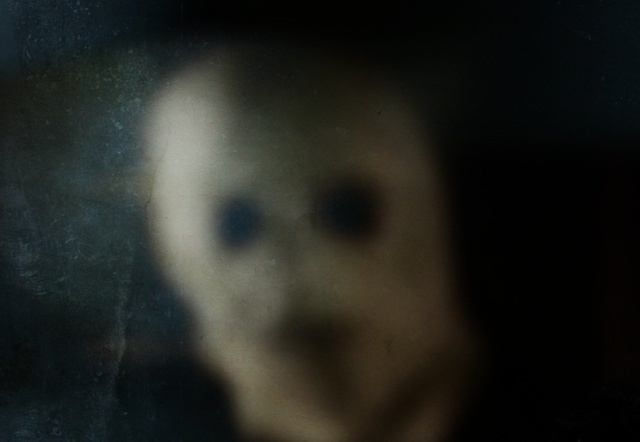}}
  \subfloat{ \includegraphics[width=1.45cm, height=1.3cm]{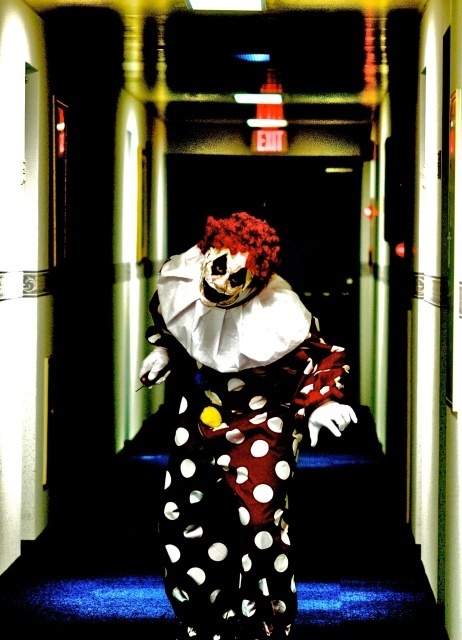}} & \subfloat{ \includegraphics[width=1.6cm, height=1.3cm]{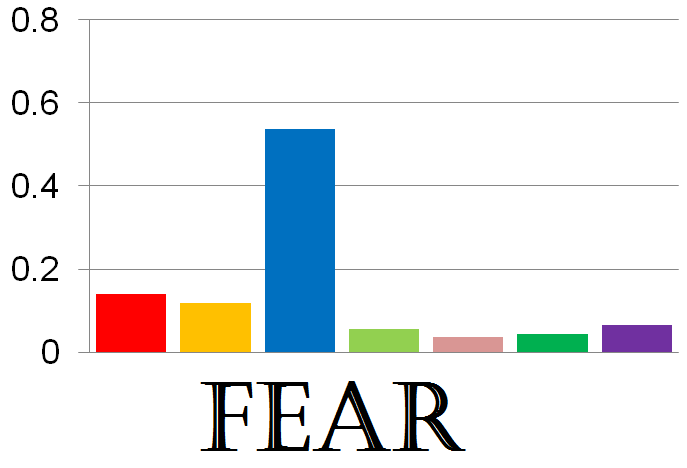}}\\ 
\cdashline{1-3}
  \small{Cemetery } & 
  \subfloat{ \includegraphics[width=1.45cm, height=1.3cm]{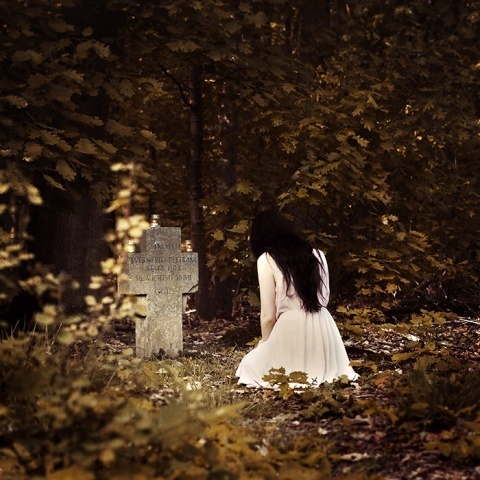}} 
  \subfloat{
  \includegraphics[width=1.45cm, height=1.3cm]{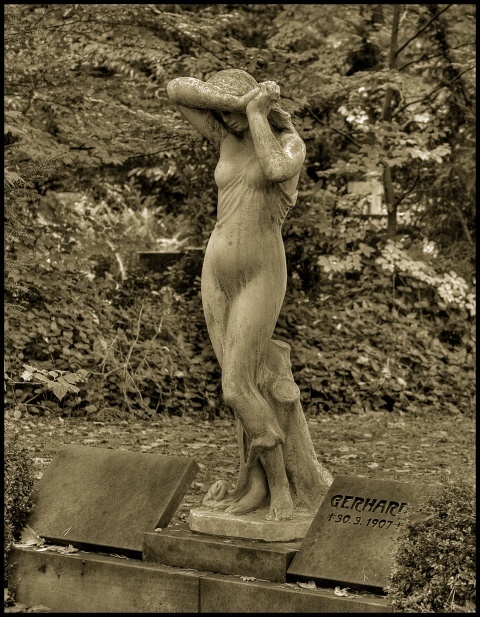}}
  \subfloat{ \includegraphics[width=1.45cm, height=1.3cm]{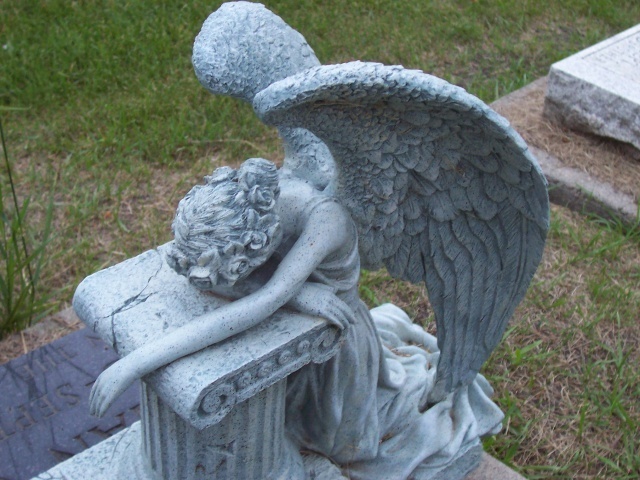}} & \subfloat{ \includegraphics[width=1.6cm, height=1.3cm]{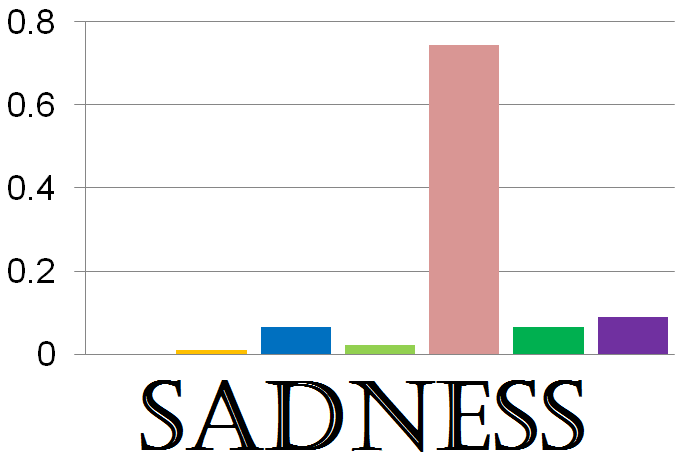}} \\
 [-1.9ex]
  \small{Lakeside }  &
    \subfloat{ \includegraphics[width=1.45cm, height=1.3cm]{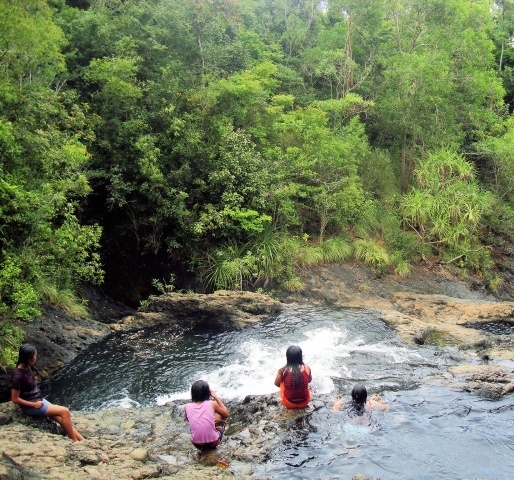}} 
  \subfloat{
  \includegraphics[width=1.45cm, height=1.3cm]{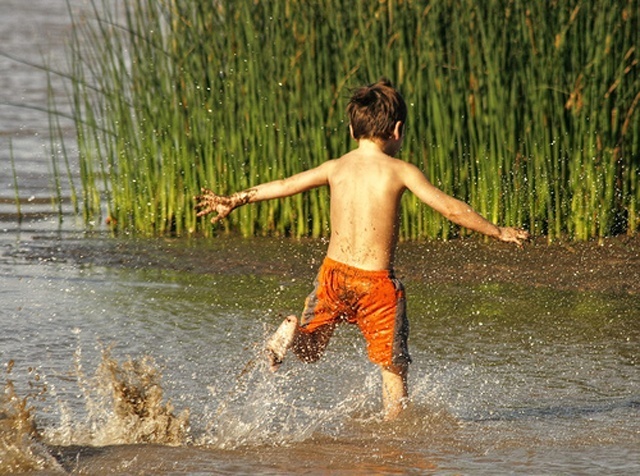}}
  \subfloat{ \includegraphics[width=1.45cm, height=1.3cm]{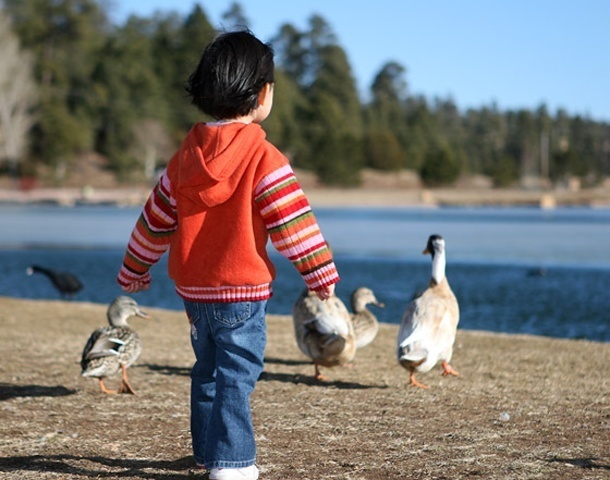}} & \subfloat{ \includegraphics[width=1.6cm, height=1.3cm]{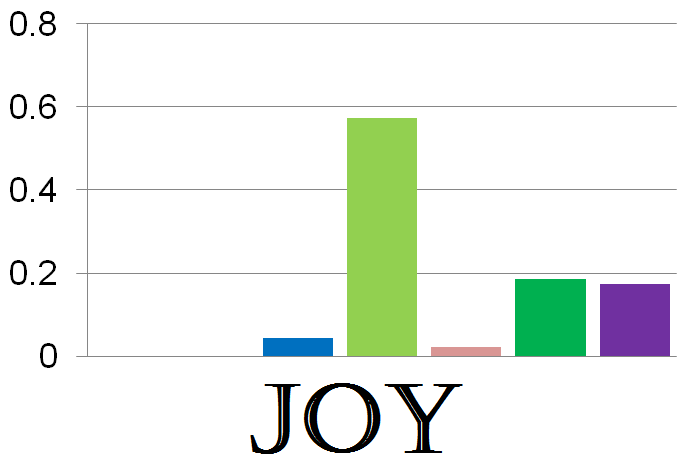}} \\
 [-1.9ex]
   \small{Garbage Dump }  & \subfloat{ \includegraphics[width=1.45cm, height=1.3cm]{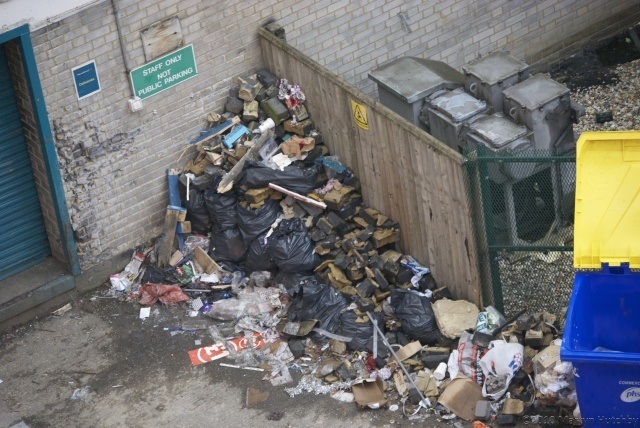}} 
  \subfloat{
  \includegraphics[width=1.45cm, height=1.3cm]{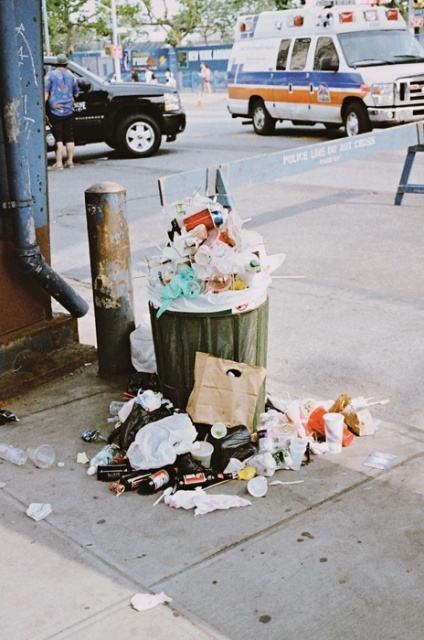}}
  \subfloat{ \includegraphics[width=1.45cm, height=1.3cm]{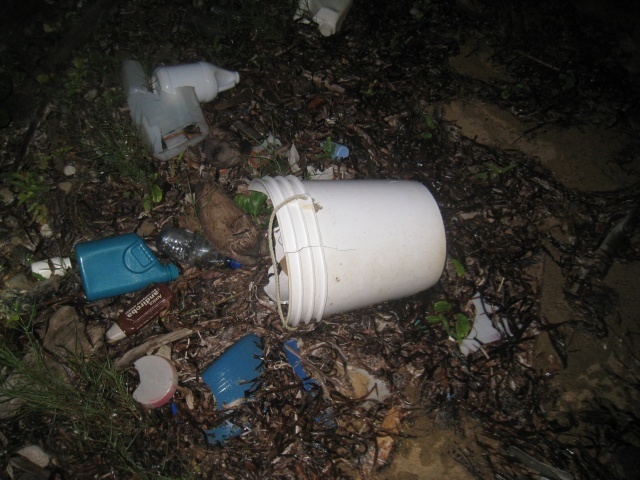}} & \subfloat{ \includegraphics[width=1.6cm, height=1.3cm]{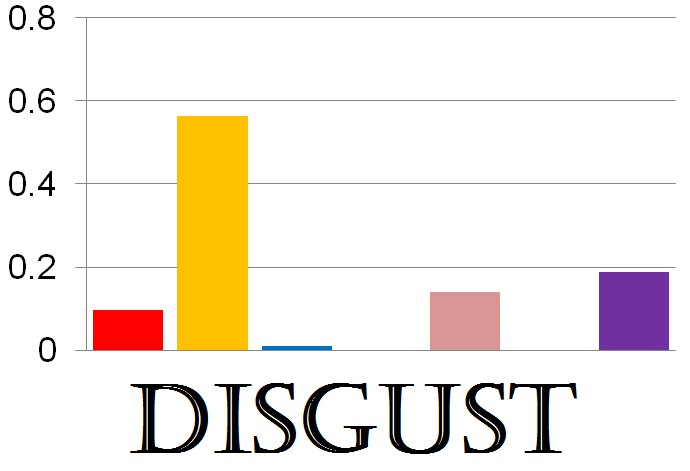}} \\
\hline
\end{tabular}
\vspace{-0.2cm}
\caption{\small{High-Level Concepts and their sample corresponding images from Emotion6 dataset with the mean emotion distribution of the displayed images shown in the right column.  Images and their respective mean emotion distribution suggest existence of strong association between specific HLCs and certain emotion classes. Color encoding scheme follows: \small{Anger \circled[anger]} \small{Disgust \circled[disgust]} \small{Fear \circled[fear] } \small{Joy \circled[joy] } \small{Sadness \circled[sadness] } \small{Surprise \circled[surprise] } \small{Neutral \circled[neutral].} }\vspace{-0.5cm}}
\label{tab:hlcsdemo}
\end{table}
\section{Introduction}

With an ever-increasing number of users sharing large quantities of multimedia data (images, videos, tweets) on social networks, there is a greater interest and need for enabling machines to interpret and relate the multimedia data like humans. Although the exact nature and process of how humans relate to this data (stimuli) is complex and remains obscure, it can be argued that there is an emotional component that plays a critical and decisive role in shaping the experience and forming the response, and this emotional component underlies the main difference between how humans and modern machines relate to their external stimuli. Therefore, bridging the gap between humans and machines requires the enabling of emotional intelligence for machines, equipping them with an ability to interact with humans on an entirely new and different level. For example, recent advances in image understanding, image retrieval and caption generation have been both rapid and substantial, e.g.
\cite{iccc2015ElgammalSaleh,he2015deep,Isola2011, Krizhevsky2012NIPSImageNet}. However, these algorithms, if described in human terms, are of a "high-IQ with a low-EQ" variety that do not interact with humans on an emotional level because of their lack of \textit{affective analysis}.  Instead of generating a list of items presented in an image, \textit{affecitve analysis} allows us to envision an artificial-intelligence system that could predict (or empathize) a viewer's emotional response to an image.  For instance, looking at the image of the magnificent Karakurum, affective analysis would enable a machine to produce more interesting responses such as \textit{Amazing view of beautiful mountains} or \textit{Scary view of treacherous heights}, instead of plainly labeling the image as "mountain". Such a capability could significantly enhance the interaction between humans and machines, and at the same time, improve a diverse array of applications such as image retrieval and caption generation.

\begin{figure}[t]
\center
 \subfloat[Surprise]{ \includegraphics[width=\textwidth, width=0.27\linewidth]{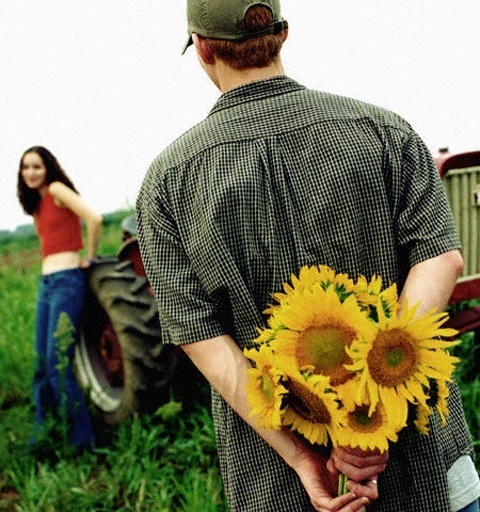} \label{fig:surprise}}
 \subfloat[Joy]     {\includegraphics[width=\textwidth, width=0.27\linewidth]{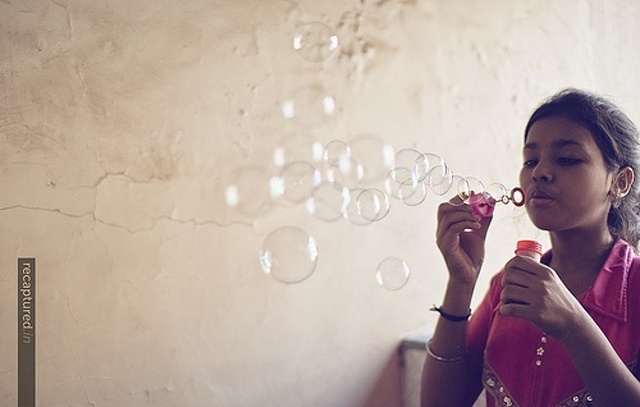}\label{fig:joy}}
  \subfloat[Fear]{ \includegraphics[width=\textwidth, width=0.27\linewidth]{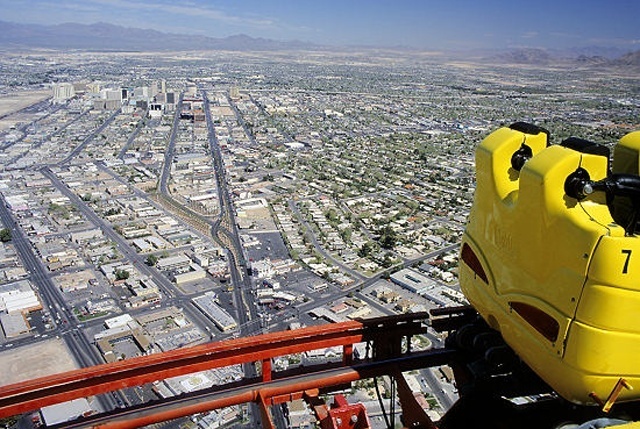} \label{fig:fear}}\\
  \center
  \vspace{-0.5cm}
 \subfloat[Amusement]{\includegraphics[width=\textwidth, width=0.27\linewidth]{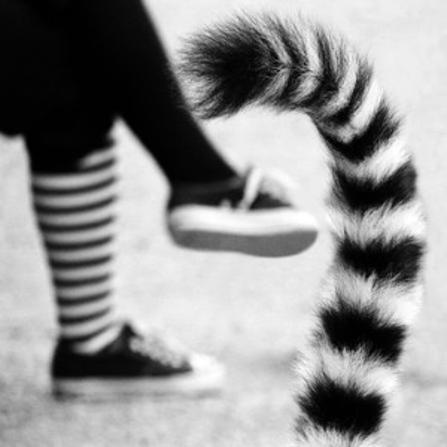}\label{fig:amusement}} 
 \subfloat[Disgust]{\includegraphics[width=\textwidth, width=0.27\linewidth]{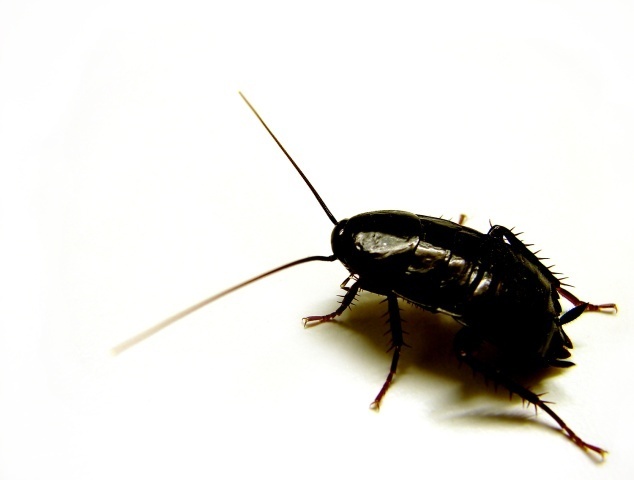}\label{fig:disgust}} 
 \subfloat[Sadness]{\includegraphics[width=\textwidth, width=0.27\linewidth]{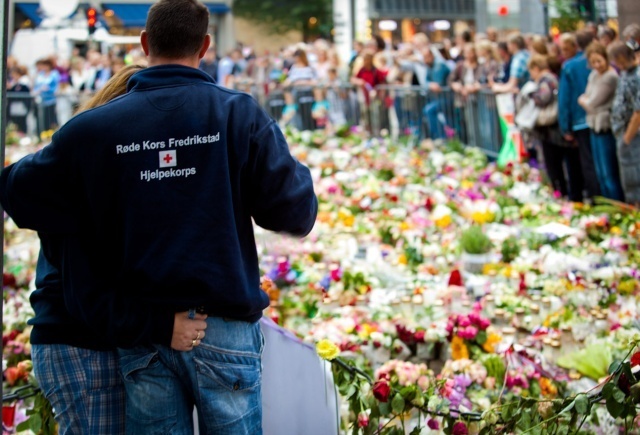}\label{fig:sadness}}
 
    \caption{\small{Sample images from Emotion6~\cite{Peng2015} and Artphoto~\cite{Machajdik2010}. The images show that low-level image features (LLF) alone are insufficient and inadequate for the understanding and modeling of viewers' emotional responses; instead, deeper concepts are required.}\vspace{-0.5cm}}\label{fig:notEmotions}
\end{figure}

As the examples in \mFIG{\ref{fig:notEmotions}} show, affective analysis of an image is a difficult problem as it is often subjective and culture-dependent. Nevertheless, there are elements such as the presence of objects and places and their relationship that are amenable to computation. For example, to understand why \mFIG{\ref{fig:joy}} should elicit joyous feeling to many viewers, one may have to detect the presence of bubbles and relate it to the presence of a child.  \mFIG{\ref{fig:surprise}~$\&$~\ref{fig:fear}} require even deeper understanding of the image content from a human's viewpoint, such as the implied notions of 'anticipation' (the hidden flower) and 'danger' (vertical height). In particular, these images also show that the typical low-level image features (LLF) would not be sufficient nor adequate to model the emotional content of an image. Due to the difficulty of reliably computing high-level image concepts in the past, a majority of the previous work on affective analysis of images have focused solely on learning the relations between various LLFs and viewer's emotional responses e.g.~\cite{wang2013interpretable, Machajdik2010}. From these work, the notion of {\it affective gap} can be observed directly as low-level image features are both incomplete and inadequate for capturing human affect.  More recently, convolutional neural networks (CNNs) have been used by~\cite{you2015robustAAAI, Peng2015} for sentiment analysis; however, CNNs have been used as a black box in these work, and without elucidation, the underlying reason for the success remains obscure. 
Since the hallmark of deep learning and convolutional neural network is their ability to capture high-level concepts, one way to interpret the results in~\cite{you2015robustAAAI, Peng2015} is the recognition of the importance of high-level concepts in affective analysis. 

More precisely, we define high-level concepts (HLCs) to be a set of attributes that provide high-level semantic and contextual information about the image: they include objects present in the image, events and places where images were taken and actions observed in the images etc.
Examples include persons, chairs, cars, flowers, sky, park, in-door/out-door and garden etc. Table.~\ref{tab:hlcsdemo} shows few high-level concepts along with corresponding images and mean emotion distributions, the images and respective mean emotion distributions indicate that there exist a relationship between affect and HLCs. 
Various recent advances in object detection and image categorization have resulted in many algorithms that can reliably extract these high-level concepts from images, and in this paper, we will leverage these advances to study affective analysis using high-level concepts in addition to the typical low-level image features. 
In particular, a collection of HLCs can be generated by applying CNNs trained for ILSVRC and Places2 competitions to each image.\ignore{ 
The main technical problem studied in this paper is the identification and utilization of the relations between computed HLCs and a set of Ekman's (basic) emotion classes~\cite{ekman1992}.}

\begin{table*}[t]
\begin{center}
\begin{tabular}{| c | c c c c c c c c c c c|} 
 \hline
\small{ Dataset} &  \small{Amusement} & \small{Anger} & \small{Awe} & \small{Contentment} & \small{Disgust} & \small{Excitement} & \small{Fear} & \small{Joy} & \small{Sadness} & \small{Surprise} & \small{Neutral} \\ [0.5ex] 
 \hline
Artphoto &
  \small{101} & \small{77} & \small{103} & \small{70} & \small{70} & \small{105} & \small{115} & \small{---} & \small{166} & \small{---} & \small{---}
 \\
Emotion6 &
  \small{---} & \small{31} & \small{---} & \small{---} & \small{245} & \small{---} & \small{329} & \small{638} & \small{308} & \small{104} & \small{325}
 \\
 \hline
\end{tabular}
\end{center}
\vspace{-0.5cm}
\caption{\small{Image Database Summary: the number of images in each emotion class in Artphoto ~\cite{Machajdik2010} and Emotion6 dataset~\cite{Peng2015}.  The two datasets share four emotion classes and they together cover eleven emotion classes.}\vspace{-0.5cm}}
\label{tab:imgProp}
\end{table*}
\vspace{-0.1cm}

Specifically, we first propose a linear admixture model that relates linearly the occurrences of HLCs to the emotional distribution vectors. Inspired by the admixture models used in document analysis (e.g. ~\cite{Blei2003}), we model the occurrences of HLCs in an image as a linear admixture associated with the Ekman's basic emotion classes.  This results in a quadratic programming problem that can be efficiently and robustly solved, yielding interesting associations between HLCs and the basic emotion classes. Examples of top HLCs extracted for each emotion class are shown in \mFIG{\ref{fig:topHLCLinear}}.
While this linear model provides reasonable descriptive power to find the associations between HLCs and various emotion classes, it lacks the predictive power to predict the emotional distribution vector of an image using the HLCs present in the image. 
This indicates that the relations between HLCs and emotion classes are unsurprisingly nonlinear, and thus motivates the development of nonlinear models in the form of support vector regression (SVR) in which kernel-based regressors are trained to predict the emotion distribution vector over basic emotion classes using HLCs as the input.
We propose a training procedure that includes a coarse feature selection component that allow each class-specific regressor to be trained using different subset of features. 
These class-specific regressors are combined into an ensemble of regressors that provides an effective and efficient predictor for predicting viewers' emotional response given an image. Experimental comparisons with existing work (e.g.~\cite{Peng2015}) show that our results are competitive, but with a clearer and more transparent view of the reason underlying its success. 

Finally, we end the introduction with a summary of the two main contributions of this paper:
\begin{itemize}
\item Instead of mapping image features directly to the emotion space\footnote{using either the low-level features or the high-level ones computed by a fine-tuned deep-network (\cite{Peng2015})}, we propose a linear admixture model for affective analysis that models the relation between HLCs and emotion classes. 
Results in \mFIG{(\ref{fig:heatMapObj}, \ref{fig:heatMapPlaces}, \ref{fig:topHLCLinear})} show that HLCs extracted are meaningful to humans and are related to different emotion classes.
\item We propose a flexible and effective approach for affective classification of images by training an ensemble of kernel-based regressors that use different subset of features for different emotion classes. For each input image, the ensemble computes a distribution over emotion classes. Experimental results reported in this paper show that the proposed approach is competitive with, and in some cases, exceed the state-of-the-art.   
\end{itemize}

\vspace{-0.1cm}
\section{Related work}
For the brevity, we briefly summarize some of the recent development in analyzing emotional content of images.
\cite{plutchik2001nature} is perhaps the most well-known seminal work that proposed the notion of wheel of emotions formed by eight primary emotions, and the seemingly unfathomable space of emotion states can then be modeled using these primary emotions as a basis. 
\cite{lang2008international} constructed the very first standard dataset for the Affective Classification problem called International Affective Picture System (IAPS). 
Subsequent work such as~\cite{Machajdik2010, Yanulevskaya2008} investigated methods and algorithms for predicting emotional response of viewers using low-level features such as brightness, contrast, saliency and texture information that are often inspired by related work in the color theory, psychology, even various theories of art and painting.
Along this line, \cite{wang2013interpretable} argued for interpretable low-level features that incorporate various aesthetic qualities for predicting emotional responses. A common shortcoming of these earlier work is that they often treat the problem as a collection of binary classification problems that assigns each image to the presence/absence of a small set of primary emotions. As the response obviously depends on the viewer, it is difficult for this kind of binary classification approach to capture the wide variation in viewer's background and culture heritage that could affect the outcome. 
More realistic modeling has been proposed in~\cite{Peng2015} that formulated the emotional response of an image as a mixture of six basic Ekman emotions~\cite{ekman1992} and a neutral one.
A new dataset named Emotion6 was also introduced by~\cite{Peng2015} in which each image has a distribution of emotional states that is used to represent various responses from different viewers.

Most of the existing work have not ventured deep into using high-level concepts for understanding emotional content of images, and the results reported in~\cite{Peng2015} using CNN (although not elaborated or discussed) strongly suggest that high-level concepts are important and crucial elements in capturing the relation between an image and the variety of emotional responses it can elicit from the viewers. Only very recently, high-level concepts have been used in the vision-to-language problem~\cite{CVPR16What} and for action classification and localization problem~\cite{JainCVPR15}. These work have shown that using explicit representations of high-level concepts can improve the state-of-art results in various computer vision problems, and our work can be considered as a reaffirmation of this observation in the context of affective computing as we will show below that high-level concepts and their relationship/associations to various emotions are also effective in modeling and predicting emotional response of viewers.

To the best of our knowledge, there has not appeared a similar work studying affective analysis using high-level concepts. Perhaps an exception can be made for~\cite{Borth2013SentiBank} and~\cite{Jiang2014EmoVid}.~\cite{Borth2013SentiBank} learned 1200 classifiers based on the adjective noun pairs (ANPs) to perform binary sentiment classification. However, relationship between ANPs and emotion classes was not explored.~\cite{Jiang2014EmoVid} combined low-level image features with high-level image features and some audio features. Their high-level concepts are formed by applying a feature fusion method to a large number of features computed using ~\cite{torresani2010Classemes},~\cite{li2014ObjectBank} and~\cite{Borth2013SentiBank}. However, the application of the feature fusion method does not allow them to harness the context in high-level semantics, since features are selected on the basis of performance of individual responses. In addition, their method of data gathering makes their work primarily about predicting emotional content in the video rather than estimating emotional response of the viewer.
Few recent works have explored the use of context to recognize facial expressions (CBFER) in the multimedia content, specifically in group settings. Rather than estimating the affect of the observer of the content, these works (\cite{mou2016automatic}, \cite{mou2015group}) deal with estimating affect expressed by the people present in the content. 

\vspace{-0.1cm}
\section{Image Datasets}
The experiments reported in this paper use the following two well-known image datasets that have appeared only in the past few years and are becoming the standard datasets for affective computing. \mTABLE{\ref{tab:imgProp}} lists the emotion classes present in each dataset and the number of images per emotion class in the given dataset.     
\vspace{-0.4cm}
\paragraph{ArtPhoto:} In ArtPhoto dataset~\cite{Machajdik2010}, there are 807 artistic photographs from eight emotion classes. These photographs were originally taken by artists with the intent of invoking a particular emotion from its viewers through appropriate use of lighting, colors and compositional parameters. Each image in the dataset has an emotion class label and in our experiments, these labels are considered as the ground truth. 
\vspace{-0.4cm}
\paragraph{Emotion6:} This dataset~\cite{Peng2015} has 1980 images, and for each image, we have a seven dimensional emotion probability distribution vector (for six Ekman's emotion classes plus an additional neutral class) and valence arousal (VA) values. These emotion probability distribution vectors were obtained through a user study, and each image is no longer associated with a single emotion class as in ArtPhoto above.  Therefore, for classification experiments, we assume the emotion class associated with each image is the one with the highest probability in the emotion distribution for that image.  

\vspace{-0.2cm}
\section{Features}
\label{sec:Features}
\vspace{-0.2cm}

Our main focus is the investigation of the relations and associations between high-level concepts and emotion classes, and in this section, we summarize the family of high-level concepts used in this paper.  Additionally, we also briefly describe the low-level features used in some of the experiments below.   

\subsection{High-Level Concepts}

In principle, we should broadly define high-level concepts to be any semantically meaningful information that can be observed from an image.  This includes both thematic and contextual information. Examples of the former include information such as the presence or absence of specific objects (e.g. persons, cars, animals) in the image that often defines the theme of the image (and the focus of the camera).  More informed thematic concepts would also include notions such as actions or events depicted by the image (e.g. football game might trigger joyous feeling while nose picking elicit the feeling of disgust).  
For the contextual information, scenes and places\ignore{(e.g. hospital, garden or beach)} where the images were taken are clearly important elements in eliciting different emotional responses from the viewers. Additional contextual information would also include out-of-image information that informs about the intention of the photographer and the historical origin of the image.   

However, in practice, due to various constraints and limitations, such a broad definition of high-level concepts is often difficult to realize, and in this paper, we do not consider out-of-image HLCs and consider only the information present in the image. Specifically, we will use a small set of concepts that include only objects and scenes, with examples including persons, chairs, cars, flowers, sky, jungle, in-door/out-door, garden, etc. This self-imposed limitation is borne out of practical necessity as the detection and identification of these concepts in the given image have been an important part of recent large scale image analysis competitions like ImageNet,  MSCOCO and Places2~\cite{zhou2014Places} and robust and stable detectors of these concepts and pre-trained classifiers are available online. 
CBFER has been not included in present set of HLC since facial expression are not visible in most of the images in ArtPhoto and images with facial expressions were removed by creators of Emotion6 dataset.
\vspace{-0.4cm}
\paragraph{Object information:} For computing the object responses, we used the AlexNet model described in~\cite{Krizhevsky2012NIPSImageNet} that has been trained on ImageNet's 1000 object categories. For each image, the CNN model outputs a $1000$-dimensional probability vector indicating the confidence value for the presence of an instance of each object category. In the following sections, we will refer to these features as \textbf{ImageNet-concepts}.
\vspace{-0.8cm}
\paragraph{Place information:} To determine the scene information present in the image, we use the model described in \cite{zhou2014Places} and has been trained on the Places205 dataset. This model outputs a $205$-dimensional vector that gives the probability of each of the $205$ places/scenes observed in the image. These features will be referred to as \textbf{Places205-concepts} in the following sections.  

Our feature vectors are simple non-negative vectors that record (probabilities of) the occurrences of these concepts in the image, and no additional information, such as the number of instances (number of times an object appears in the image), spatial location of the object or its size, are used. As will be shown by the experiments below, these simple HLCs feature vectors are able to capture sufficient amount information for affective analysis.
\vspace{-0.1cm}
\subsection{Low-Level Features}
In addition to these HLCs, low-level image features related to texture, composition, saliency, color and edge information are also computed as described in~\cite{Peng2015}, resulting in a 628-dimensional feature vector. These features have been normalized so that their values are all between $0$ and $1$.
\vspace{-0.1cm}
\section{A Linear Admixture Model for HLCs and Emotion distributions}
\label{sec:subspace}
\noindent 
\begin{figure*}[t]
\center
\subfloat[HLC from imageNet]{ \includegraphics[width=0.9\textwidth]{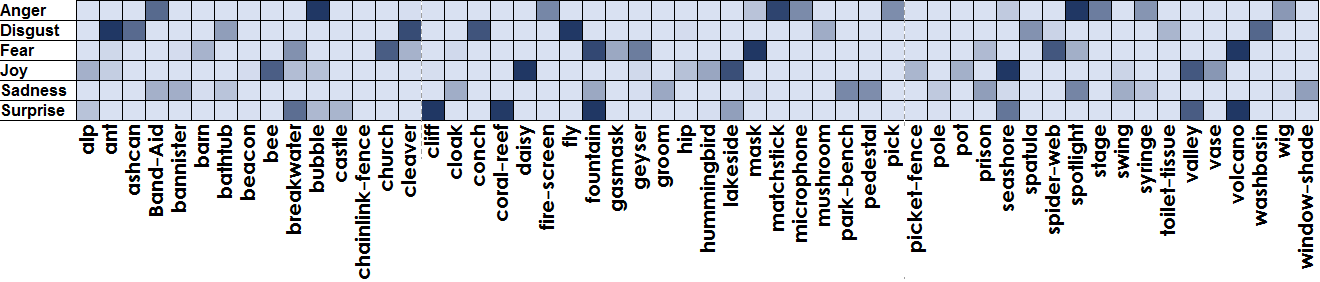} 
\label{fig:heatMapObj}\vspace{-0.7cm}}\\\vspace{-0.3cm}
\subfloat[HLC from Places205]{ \includegraphics[width=0.9\textwidth]{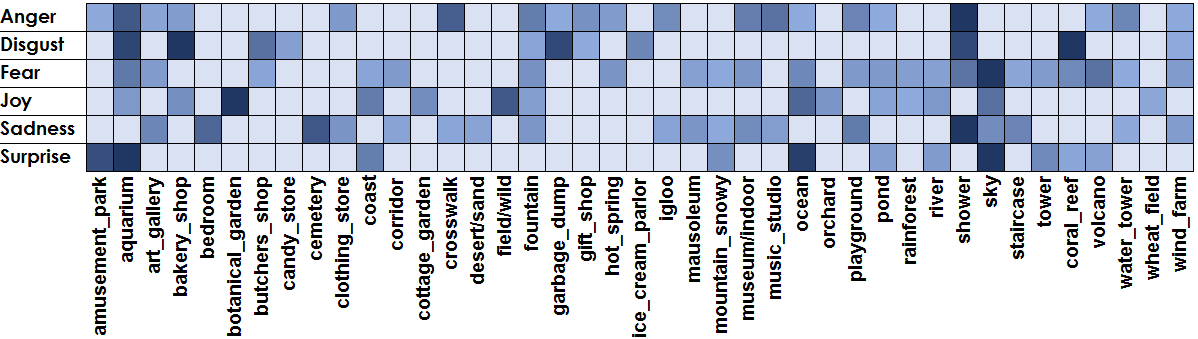} 
\label{fig:heatMapPlaces}}
\caption{\small{Heat map showing the association between HLCs (concepts in ImageNet and Places2015) and the six emotion classes. These values ($e_i$) are calculated using Eq.(\ref{eq:subsp}). Higher values of $e_i$ are shown in daker shades.}\vspace{-0.3cm}}
\end{figure*}
Inspired by the earlier work in document/topic analysis~\cite{Blei2003}, we propose a linear admixture model to model the relation between HLCs and emotion distribution. Recall that in document/topic analysis, the word count in each document is assumed to be an admixture of a set of topics, with each topic specifying a particular word distribution. In our context, the words are replaced by HLCs and the emotion states take the place of topics. More importantly, the word count now corresponds to the probability of occurrence of a given HLC in the image. More specifically, each image is associated with a $d$-dimensional vector $\bh$, HLC distribution vector, where $d$ is the feature dimension (the number of HLCs)\ignore{and each non-negative component of $\bh$ gives the confidence for the appearance of each HLC in the image}. In practice,  this vector is determined by the (softmax) output of the last layer of neuron in a convolutional neural network applied to the image.  The $\bh$ is normalized so that its components sum to one, and we interpret each component of $\bh$ as the occurrence probability of the corresponding HLC in the image.  In our proposed admixture model,  it is assumed that the vector $\bh$ is a linear combination: $\bh = e_1 \bp_1 +e_2 \bp_2 + ... + e_7 \bp_7$, where $e_1, ..., e_7$ denote the seven components of the emotion distribution vector with $e_1, ..., e_7 \geq 0, e_1+e_2+...+e_7=1$, and $\bp_1, ..., \bp_7$ are the HLC distribution vectors correspond to the seven emotion classes.        
\begin{figure*}[th]
\includegraphics[width=1.0\textwidth]{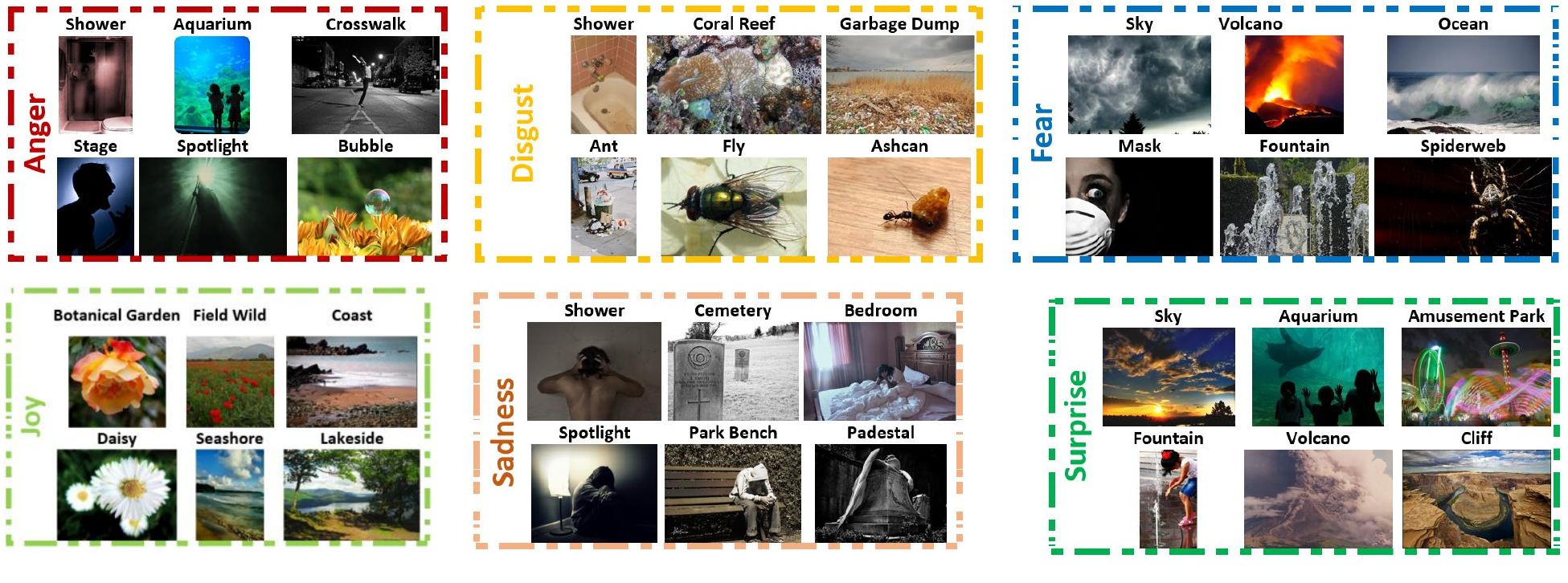} 
\vspace{-0.9cm}
\caption{\small{Top high-level concepts (HLC) selected for each emotional category for Emotion6 dataset~\cite{Peng2015} using Eq.(\ref{eq:subsp})}\vspace{-0.5cm} }
\label{fig:topHLCLinear}
\end{figure*}
For affective analysis, we are given a collection of (training) images with known emotion distribution vector.  For each training image, we can also obtain its HLC distribution vector, and the aim of the analysis is to determine the HLC distribution vectors associated with the seven emotion classes.  With the linear admixture model above, the seven HLC distribution vectors can be determined using quadratic programming as follows.\ignore{  For each image, let $e_1, ..., e_7$ denote the seven components of its emotion state vector.  We assume the following linear relation holds: $\bh = e_1 \bp_1 +e_2 \bp_2 + ... + e_7 \bp_7.$}

Let $\bx=[\bp_1; \bp_2; ...; \bp_7]$ be a vertical concatenation of probability vectors and 
$\bA$ be a $d\times 7d$ matrix structured as $7$ horizontal blocks: 
$\bA = [\,\, e_1 \bI_{d\times d}\,\, e_2 \bI_{d\times d}\,\, ...
e_7 \bI_{d\times d}\,\,]$,  
where $\bI_{d\times d}$ is the identity matrix.  Clearly, $\bx\in \mathbb {R}^{7d}$, and we can rewrite the above equation using matrix $\bA$ as $\bh=\bA \bx$.  
For $n$ images, we define a constrained optimization problem using the objective function $\min \sum_{i=1}^n \| \bh_i - \bA_i \bx\|^2 $
subject to the constraints that components of $\bp_i$ are
non-negative and sum to one.

The last equation can be put into
the standard form $\min \| \bY - \bA \bx\|^2$
where $\bY=[\bh_1;\bh_2; ...;\bh_n]$ and $\bA=[\bA_1; \bA_2; ...;
\bA_n]$.  Multiplying out the above equation we have
\vspace{-0.1cm}
\begin{equation}
\label{eq:subsp}
\begin{aligned}
& \underset{x}{\text{min}}
& & \bx^{\top}\bA^{\top}\bA \bx - 2\bY^{\top}\bA\bx +
\bY^{\top}\bY \\
& \text{s.t}
& & \bC \bx = 1_7,
\end{aligned}
\end{equation}
where $\bC = \text{kron}(\text{eye}(7), \text{ones}(1, d))$ is the Kronecker product between a $7\times 7$ identity matrix with $1_7$, a seven-dimensional column vector of ones. $\bC \bx = 1_7$ represent the constraint that $\sum_{j=1}^{d} \bp_i^j = 1,  \forall i \in \{1, ..., 7\}$.
This is the standard quadratic programming problem with linear constraints, and can be readily solved using several different algorithms.
\subsection{Qualitative Experiment}
The matrix $\bx=[\bp_1; \bp_2; ...; \bp_7]$ can be directly learned from the dataset using Eq.(1). For each training image, its corresponding $\bh$ is set to \textbf{ImageNet-concepts} extracted from the image as \textbf{HLC}, and additionally, we only used the features that were detected at least $10\%$ of the time. By solving the quadratic programming problem, we can obtain the matrix $\bx$ and each of its columns provides a HLC probability distribution for one emotion class.  To show how concepts differ across the emotion classes, we select top ten concepts (ten largest components in $\bp_i$) for each emotion class and take their union.  The heat map in \mFIG{\ref{fig:heatMapObj}} shows how certain concepts are important, some appear (with high value) in multiple categories while others appear only in one.  More importantly, the associations found by the proposed method agree reasonably well with our general expectation on how various HLCs should be associated with particular emotion classes.    

We repeat the experiment using \textbf{Places205-concepts}. The learned heat map is shown in \mFIG{\ref{fig:heatMapPlaces}}.
These results corroborate well with the earlier results using ImageNet-concepts.
In Places205-concepts, \textit{Disgust}-emotion gives high value to 'garbage-dump' and 'shower' while from imageNet-concepts it picks out 'ant', 'ashcan', 'bathtub', 'fly' and 'washbasin'.
Joy on the other hand is represented by the concepts related to flowers, greenery and open-spaces ('daisy', 'lakeside', 'wheat-field', 'orchard'). Six of the top-concepts (having high scores) with respective images (from Emotion6 dataset) are shown in \mFIG{\ref{fig:topHLCLinear}}.  There are some concepts that appear important for more than one category, for example, 'rainforest' appears in both \textit{Joy} and \textit{Fear}, 'corridor' appears in \textit{Fear} and \textit{Sadness}.  While the linear admixture model is able to detect the association between each emotion class and multiple HLCs, the dependency structure as revealed by the two heat maps suggests that the exact relations between them are far from linear as expected.  In particular, this justifies the choice of non-linear kernel in the following section for the affective classification problem. 


\vspace{-0.3cm}
\section{Experiments}
\vspace{-0.2cm}
Experiments are performed on Artphoto \cite{Machajdik2010} and Emotion6 \cite{Peng2015} dataset, with primary interest in Emotion6 dataset since it, unlike Artphoto, has emotion distribution for each image, making it closer to reality. We follow same protocol for experiments as used by \cite{Peng2015}. \textcolorM{blue}{Due to high class imbalance in the Emotion6 dataset, training classifiers is quite challenging, instead we perform support vector regression (SVR) on the probability of each emotion.\ignore{ Predicted class is chosen by picking up the emotion class which has the highest probability.}}

\begin{table}[h]
\center
\begin{tabular}{ |l | c | c c|}
\hline
\textbf{Features}	&  \small{\textbf{Acc. ($\%$)}} &\small{\textbf{KLD}}	& \small{\textbf{BC}}\\
\hline
\small{ LLF \cite{Peng2015} } & \small{ 38.9 } & \small{ 0.577 } & \small{ 0.820 } \\
\small{CNNR \cite{Peng2015}} & --- &\small{ 0.480 } & \small{ 0.847 } \\
\hline
\small{ImageNet} & \small{ 45.83 } & \small{ 0.559} & \small{ 0.818 } \\
\small{Places205} & \small{ 43.16 } & \small{ 0.548 } & \small{ 0.827 }\\
\small{LLF, ImageNet} & \small{ 44.50 } & \small{ 0.588} & \small{ 0.824 } \\
\small{LLF, Places205} & \small{ 47.00 } & \small{ 0.583} & \small{ 0.831 } \\
\small{ImageNet, Places205} & \small{ 49.83 } & \small{ 0.518} & \small{ 0.830 } \\
\small{LLF, ImageNet, Places205} & \small{ 42.00 } & \small{ 0.574} & \small{ 0.823 } \\
\textbf{Hybrid Model } & 52.00  &  0.493  & 0.839  \\
\hline
\end{tabular}
\caption{\small{Accuracy of SVR-based classifiers for Emotion6 dataset trained on different sets of low level and high level features.}\vspace{-0.4cm}}
\label{tab:classAccuracy}
\end{table}
   \textcolorM{blue}{We train seven SVRs with RBF kernel, one for each emotion class,  using the given feature subset (Sec.~\ref{sec:Features}). During training phase, SVR parameters ($C$ and $\gamma$) are optimized through a grid search using 5-fold cross validation (only on the training set) with objective function being minimization of Mean Square Error (MSE). In the testing phase SVR machine for each emotion class outputs probability of that emotion for the image under consideration. Each negative element of this seven dimensional output is clipped to zero and then normalization is applied so that vector sums up to 1, hence generating a predicted emotion distribution.}
    
\textcolorM{blue}{Accuracy of the distribution prediction is measured by comparing predicted distribution with ground truth, using KL-Divergence (KLD) and Bhattacharya coefficient (BC). The \textit{classification accuracy} is calculated by checking if most dominant emotion both in predicted distribution and in the ground truth distribution are same. 
Our results (\mTABLE{\ref{tab:classAccuracy}}) on training SVR using fusion of different HLC and LLF (where same feature-vector is used to train SVR for each emotions), exhibited classification accuracy better than \cite{Peng2015}. However, our predicted distribution's quality was not comparable.}

\subsection{Hybrid model} \label{hybrid model}
\label{sec:hybrid}
\begin{figure}[th]
 \centerline{
 \includegraphics[height= 5.0cm, width=1\linewidth]{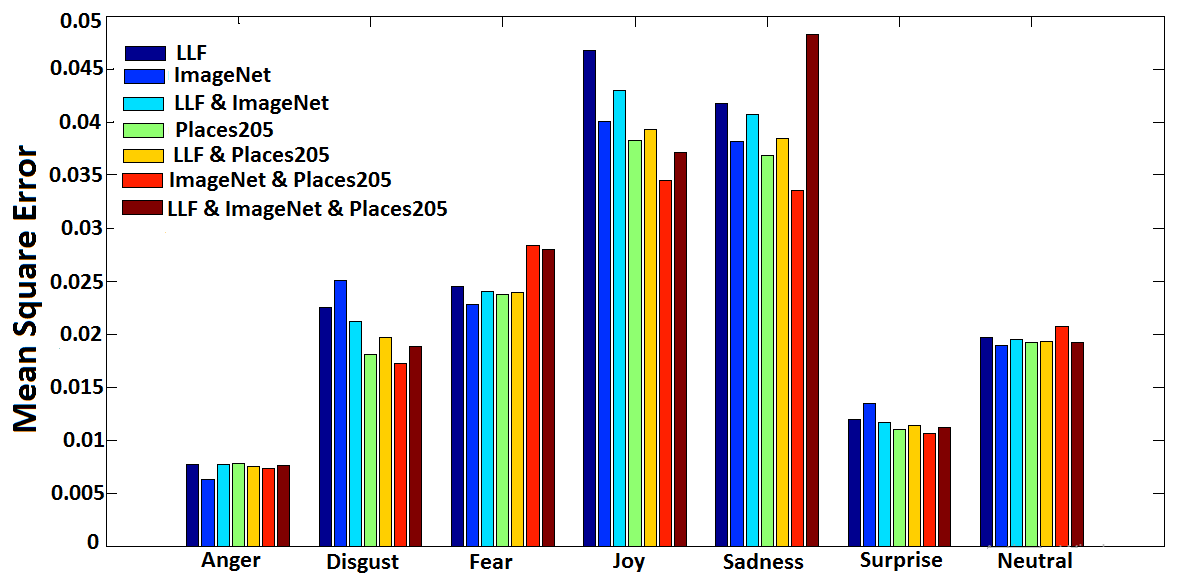} 
 }
 \vspace{-0.2cm}
 \caption{\small{Training MSE for each class in Emotion6 dataset using different features subsets. The results clearly shows that different emotions require different feature subset for classification.} \vspace{-0.2cm}}
\label{fig:svrError}
\end{figure}

\vspace{-0.2cm}
To test our hypothesis that each emotion class might be relying on the different set of concepts and high-dimensionality of these concepts with comparably smaller training set is not allowing it to learn that relationship, we designed algorithm to create ensemble of regressors. For each emotion in Emotion6 dataset we train seven classifiers (or regressors), each on different subset of features. For each emotion class, regressor having smallest training mean square error (MSE) is picked to create ensemble. Fig.\ref{fig:svrError} shows training MSE for each SVR trained for Emotion6 dataset and \mTABLE{\ref{tab:hybrid}} lists the final feature subset selected for each emotion category in Emotion6 to create ensemble. This ensemble of regressors, each trained on different feature set is called \textbf{Hybrid Model (HM)}. 
\begin{table}[th]
\center
\begin{tabular}{| m{4.5em} | m{2.8cm}| m{2.4cm}|}
\hline
\textbf{Emotion}	& \textbf{Emotion6}  & \textbf{Artphoto} \\
\hline
\small{Anger} 		& \small{ImageNet}	& \small{\PLACES }\\
\hline
\small{Sadness} 	& \small{ImageNet+\PLACES}	& \small{LLF+ImageNet}\\
\hline
\small{Fear} 		& \small{ImageNet}			& \small{\PLACES }\\
\hline
\small{Disgust} 	& \small{ImageNet+\PLACES}	& \small{LLF} \\
\hline
\small{Joy}			& \small{ImageNet+\PLACES}	&\small{--} \\
\hline
\small{Surprise} 	& \small{ImageNet+\PLACES} 	& \small{--} \\
\hline
\small{Neutral}	 	& \small{ImageNet} 			& \small{--} \\
\hline
\small{Amusement} 	& --  						& \small{LLF+ImageNet+\PLACES} \\
\hline
\small{Excitement} 	& -- 						& \small{ImageNet+\PLACES} \\
\hline
\small{Contentment} & -- 						& \small{LLF} \\
\hline
\small{Awe} 		& -- 						& \small{LLF+ImageNet+\PLACES} \\
\hline
\end{tabular}
\caption{\small{Features selected for each emotion on the basis of training MSE for Emotions6 \cite{Peng2015} and Artphoto dataset \cite{Machajdik2010} to create HM. "--" indicates emotional class not presented in that dataset.}\vspace{-0.5cm}}
\label{tab:hybrid}
\end{table} 

\textcolorM{red}{Using \textbf{HM} on Emotion6 dataset we see considerable improvement in both classification-accuracy and distribution-accuracy, comparing to when only one set of the features were used for all emotions \mTABLE{\ref{tab:classAccuracy}}. Quality of our predicted emotion distribution is closer to \cite{Peng2015} and our classification accuracy surpasses SVR trained on only low level features in  \cite{Peng2015}, with considerable margin (they do not share their classification accuracy results for CNNR method)\footnote{It should be noted that our method is using output of CNN trained on imagenet data, where as \cite{Peng2015} used CNN that were fine-tuned on the Emotion6 dataset. However, information we are generating is more helpful than simple output of emotional distribution. Our method is connecting emotions with tangible and understandable concepts.}. We also observed that errors in distribution accuracy were made where distribution entropy is high, indicating inherent ambiguity in the distributions where every class has similar or almost similar probability, also confuses the SVR.}

\begin{figure}[h]
 \centerline{
 \includegraphics[width=0.9\linewidth, height= 5.5cm]{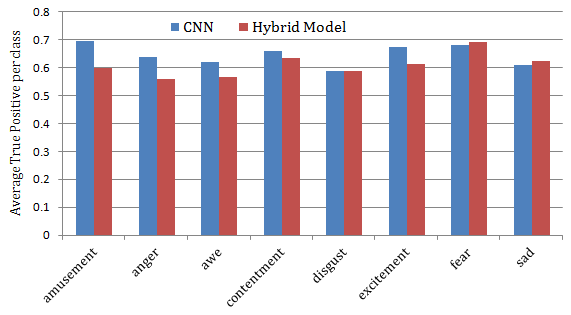}
 \vspace{-0.5cm}
 }
 \caption{\small{Classification performance of hybrid model and CNN \cite{Peng2015} with ArtPhoto dataset.}\vspace{-0.2cm}}
\label{fig:artPhotoTP}
\end{figure}

For ArtPhoto dataset, we train seven SVM classifiers, in similar fashion as above, for each of 8 emotions under 1-vs-all setting. In order to minimize class biasness during training, similar to ~\cite{Peng2015}, we repeat the positive examples such that number of positive and negative examples are same. 5-fold cross validation was used to evaluate the results.
After training these classifiers, ensemble was created by picking up the one classifier for each emotion that has maximum average true positive rate for training.
As images in Artphoto dataset are purposely taken (with appropriate use of lightning and other parameters) to invoke specific emotion in its viewers, therefore, LLF like colors and textures played an important role during classification. This can be seen from \mTABLE{\ref{tab:hybrid}}, where in 5 out of 8 emotion categories low level features are chosen to create ensemble for Artphoto dataset.
As shown in Fig. \ref{fig:artPhotoTP} our average true positive rate per class for Artphoto dataset is comparable to the one that has been computed using CNN~\cite{Peng2015}.
\ignore{
\begin{figure}
 \centerline{
 \includegraphics[width=1\linewidth]{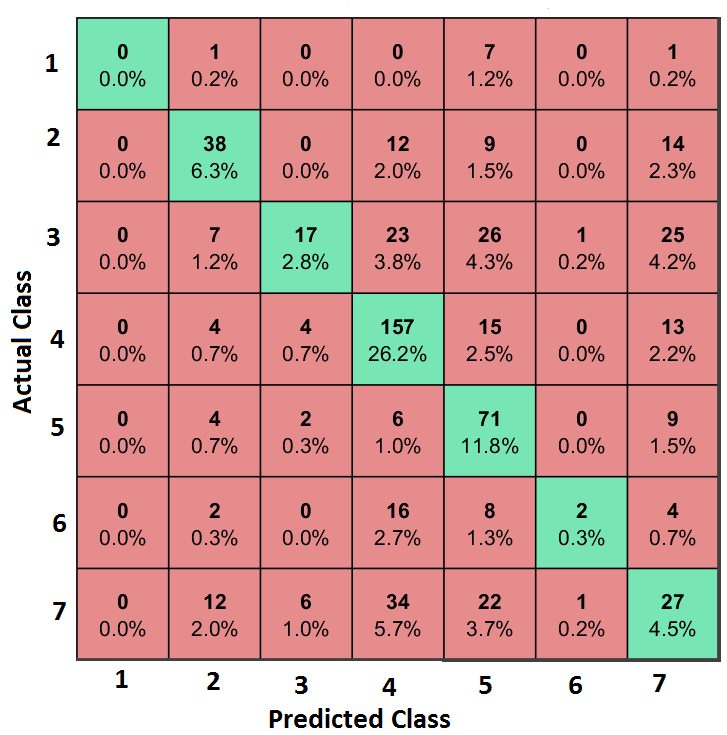}
 }
 \caption{\small{Classification confusion matrix. On horizontal axis we have actual class numbers. Here 1 is used for anger, 2 for disgust, 3: fear, 4: joy, 5: sadness, 6: surprise and 7:neutral}}
\label{fig:confMat}
\end{figure}}
\vspace{-0.3cm}
\subsection{VA prediction using hybrid model}
\textcolorM{red}{There exist different models for emotions, some represent affect using discrete categories (e.g. \cite{plutchik2001nature} and \cite{ekman1992}) while other map affect in a multi-dimensional space. One such model is Circumplex model~\cite{russell1980} that maps affect in 2-D space using valence and arousal (VA) values where valence represent pleasantness of the emotion and arousal represent its strength.
Both categorical and dimensional models cannot totally replace each other, therefore, we have also trained a VA hybrid model to predict valence and arousal values for Emotion6 dataset. Same experiment settings were used as described in Sec.\ref{hybrid model}, except we regress over valence/arousal score instead of emotion probability values. We used average of absolute difference (AAD) to evaluate how good our results are with respect to ground truth; lower the AAD, better would be our predictions. \mTABLE{\ref{tab:vaScore}} lists our VA results using different feature sets. These results show that even for VA prediction our hybrid model outperforms all other regressors, which have been trained on individual feature sets. Our hybrid model even outperforms CNNR and SVR results in ~\cite{Peng2015} which were mapping low level image-features to these values, where as we have mapped HLCs to Valance and Arousal.}
 \begin{table}[h]
\center
\begin{tabular}{ |l | c c|}
\hline
\textbf{Features}	& \small{\textbf{Valence AAD}}	& \small{\textbf{Arousal AAD}}\\
\hline
\small{ LLF \cite{Peng2015} } & \small{ 1.347 } & \small{ 0.734 } \\
\small{CNNR \cite{Peng2015}}  &\small{ 1.219 } & \small{ 0.741 } \\
\hline
\small{ImageNet} & \small{ 1.5851} & \small{ 0.6898 } \\
\small{Places205} & \small{ 1.3544 } & \small{ 0.6826 }\\
\small{LLF,ImageNet} & \small{ 1.5831} & \small{ 0.6781 } \\
\small{LLF,Places205} & \small{ 1.2093} & \small{ 0.6766 } \\
\small{ImageNet,Places205} & \small{ 1.273} & \small{ 0.6802 } \\
\small{LLF,ImageNet,Places205} & \small{ 1.265} & \small{ 0.6703 } \\
\textbf{Hybrid Model } & \textbf{ 1.2093 } & \textbf{ 0.6802 }\\
\hline
\end{tabular}
\caption{\small{Emotion6 VA scores based on SVR-based classifiers trained on different HLC.}\vspace{-0.8cm}}
\label{tab:vaScore}
\end{table}

\section{Conclusion}

In this paper, we have explored the relations/associations between high-level concepts and emotion classes. 
In particular, we have found that, for affective classifications, different emotion classes require different sets of HLCs, and the best classification results are achieved when both the high-level concepts and low-level features are utilized. Accordingly, we have proposed a class-dependent feature fusion (Hybrid Model) method that automatically chooses the required set of concepts (e.g. objects or place or both) for training, and the proposed  
emotion (probability) prediction algorithm achieves results that are similar to the state-of-the-art.

Unlike previous work that, conceptually, map image features directly to emotion space, we propose to map HLCs to the space of emotion distributions. Experimental results have shown that, at least for affective classification problem, our results are comparable to the state-of-the-art results~\cite{Peng2015}. In addition to this, our approach offers two important advantages.  First, mapping HLCs to emotion distributions allows us to disassociate affective computation from the image feature extraction step. In particular, HLCs could be obtained from different sources other than image processing such as image caption, and this allows for a shorter training time with more precise and diverse training inputs. Second, an important advantage of using HLCs is that they can be easily recognized by humans and therefore, it can be more readily applied to tasks such as image retrieval and caption generation~\cite{CVPR16What} with an affective component.
    
Experimentally, we have demonstrated that using HLCs (with LLFs), we can predict dominant emotion response for an image with $52\%$ accuracy, better than any previous result reported on Emotion6 dataset~\cite{Peng2015}. Additionally, using LLFs and HLCs, we are able to obtain the state-of-art results on predicting VA-scores of Emotion6 dataset shown in Table.~\ref{tab:vaScore}. We remark that as we have used pre-trained classifiers to obtain the concept probabilities, our HLC-detecton step is dependent upon the accuracy of these pre-trained classifiers. While these classifiers are generally efficient and reliable, there are instances where incorrect classification results are produced. However, our experimental results have shown that even with these errors, our method is still able to achieve comparable performance with the state-of-the-art methods.   

In this paper, we have used only objects and places as the high-level concepts. For future work, we plan to substantially enlarge the scope of HLCs to include more diverse and better-informed information such as human actions, pose and facial expressions of people in the image. 

{\small
\bibliographystyle{ieee}
\bibliography{egbib}
}

\end{document}